\pdfoutput=1
\documentclass[11pt]{article}

\usepackage{EMNLP2023}

\usepackage{times}
\usepackage{latexsym}

\usepackage[T1]{fontenc}
\usepackage[utf8]{inputenc}

\usepackage{microtype}

\usepackage{inconsolata}

\usepackage{url}
\usepackage{wrapfig}
\usepackage{booktabs}
\usepackage{multirow}
\usepackage{fdsymbol}
\usepackage{rotating}
\usepackage{caption}
\usepackage{subcaption}

\title{Embedding Structure Matters: Comparing Methods to Adapt Multilingual Vocabularies to New Languages}

\author{
    C.M. Downey$^{\alpha}$ \quad
	Terra Blevins$^{\beta}$ \quad
        Nora Goldfine$^{\alpha}$ \quad
    Shane Steinert-Threlkeld$^{\alpha}$\\
    $^\alpha$Department of Linguistics, University of Washington \\
	$^{\beta}$Paul G. Allen School of Computer Science \& Engineering, University of Washington \\
	{\tt \{cmdowney,shanest\}@uw.edu} \\
    {\tt blvns@cs.washington.edu} \\
    {\tt ngoldfine@gmail.com}
}

\begin{document}
\maketitle
\begin{abstract}
Pre-trained multilingual language models underpin a large portion of modern NLP tools outside of English. A strong baseline for specializing these models for specific languages is Language-Adaptive Pre-Training (\textsc{Lapt}). However, retaining a large cross-lingual vocabulary and embedding matrix comes at considerable excess computational cost during adaptation. In this study, we propose several simple techniques to replace a cross-lingual vocabulary with a compact, language-specific one. Namely, we address strategies for re-initializing the token embedding matrix after vocabulary specialization. We then provide a systematic experimental comparison of our techniques, in addition to the recently-proposed \textsc{Focus} method. We demonstrate that: 1) Embedding-replacement techniques in the monolingual transfer literature are inadequate for adapting multilingual models. 2) Replacing cross-lingual vocabularies with smaller specialized ones provides an efficient method to improve performance in low-resource languages. 3) Simple embedding re-initialization techniques based on script-wise sub-distributions rival techniques such as \textsc{Focus}, which rely on similarity scores obtained from an auxiliary model.
\end{abstract}

\section{Introduction}
\label{sec:intro}

For languages other than English and a handful of other very high-resource languages, pre-trained multilingual language models form the backbone of most current NLP systems. These models address the relative data scarcity in most non-English languages by pooling text data across many languages to train a single model that (in theory) covers all training languages \citep[i.a.]{devlin-2019-mbert, conneau-lample-2019-cross, conneau-etal-2020-unsupervised, liu-etal-2020-multilingual-denoising, workshop2023bloom}. These models often include language-agnostic tokenization and an increased vocabulary capacity over monolingual models \cite{conneau-etal-2020-unsupervised}.  

However, \citet{wu-dredze-2020-languages} show that these massively multilingual models still underperform on lower-resource languages. Recent efforts to cover these languages instead pre-train models that are specialized to specific languages or language families \cite{ogueji-etal-2021-small, ogunremi-etal-2023-mini}. These approaches nonetheless require training a new model from scratch and do not leverage transferable information in existing models.

Our study builds on a line of work which instead \textit{adapts} a pre-trained cross-lingual model (such as XLM-R; \citealp{conneau-etal-2020-unsupervised}) to a single language, or a smaller set of languages. Language-Adaptive Pre-Training (\textsc{Lapt})---continuing the MLM or CLM pre-training task on only the target language(s)---is a simple and strong baseline in this regard \citep{chau-etal-2020-parsing}.

However, \textsc{Lapt} with no change to the cross-lingual vocabulary comes with considerable excess computational cost: when adapting to a single language or small subset of languages, only a small fraction of the cross-lingual vocabulary is used. The excess vocabulary still contributes to the computational cost on both the forward and backward pass, and embedding/output matrices often constitute a large fraction of the total trainable model parameters (for XLM-R-base, 192M / 278M $\approx$ 69\% of parameters). Additionally, the information-theoretic tokenization modules for cross-lingual models are usually under-optimized for any given language, and especially low-resource languages \citep[i.a.]{acs-2019-exploring, conneau-lample-2019-cross}

For this reason, we propose several simple techniques to replace the large cross-lingual vocabulary of a pre-trained model with a compact, language-specific one during model specialization. Training a new SentencePiece or BPE tokenizer poses no special difficulties. However, re-initializing the embedding matrix for a new vocabulary, which will almost certainly introduce many new tokens lacking pre-trained embeddings, poses significant challenges.  We compare several methods for such embedding re-initialization.

After reviewing related literature in Section~\ref{sec:related_work}, we conduct a qualitative exploration of the pre-trained embedding space for a standard multilingual model: XLM-R (Section~\ref{sec:embedding_exploration}). This exploration informs our formalization of simple techniques to align new vocabulary embeddings with the pre-trained embedding distribution of our base model (Section~\ref{sec:reinit_techniques}). We then provide a systematic experimental comparison of the embedding re-initialization techniques we propose, plus the recently proposed \textsc{Focus} re-initialization method \citep[Section~\ref{sec:experiments}]{dobler-de-melo-2023-focus}. Our experiments cover a wide selection of low- and mid-resource target languages (i.e. those that have the most to gain from language specialization).\footnote{The software used to run all experiments may be found at \url{https://github.com/cmdowney88/EmbeddingStructure}}

The results of our experiments (Sections~\ref{sec:results}, \ref{sec:discussion}) demonstrate the following: 1) Embedding-replacement techniques proposed in the monolingual model adaptation literature are inadequate for adapting multilingual models. 2) Replacing large cross-lingual vocabularies with smaller language-specific ones provides a computationally-efficient method to improve task performance in low-resource languages. 3) The simple re-initialization techniques we propose here, based on script-wise embedding sub-distributions, rival techniques such as \textsc{Focus}, which rely on model-driven semantic similarity.

\section{Related Work}
\label{sec:related_work}

\paragraph{Pre-trained Model Adaptation}
Extensive work has proposed re-using and modifying pre-trained models for new settings in order to retain existing model knowledge and reduce pre-training costs. \citet{gururangan-etal-2020-dont} show that continued training on domain-specific data effectively adapts pre-trained models to new domains in both high- and low-resource settings. This approach is also used to adapt models to new languages (i.e. Language-Adaptive Pre-Training / \textsc{Lapt}; \citealp{chau-etal-2020-parsing}).

Other approaches involve training new, language-specific adapter layers to augment a frozen monolingual \cite{artetxe-etal-2020-cross} or multilingual encoder \cite{pfeiffer-etal-2020-mad, ustun-etal-2020-udapter, faisal-anastasopoulos-2022-phylogeny}. A comparison of these cross-lingual adaptation approaches \cite{ebrahimi-kann-2021-adapt} found that continued pre-training often outperforms more complex setups, even in low-resource settings. With this in mind, our experiments evaluate the success of models tuned for target languages with \textsc{Lapt}, starting from variable initializations depending on a choice of embedding adaptation technique.

\paragraph{Cross-lingual Vocabulary Adaptation}

A major limitation in adapting pre-trained models to new languages is the subword vocabulary, which often fails to cover an unseen script \cite{pfeiffer-etal-2021-unks} or tokenizes target text inefficiently \cite{acs-2019-exploring}. \citet{muller-etal-2021-unseen} demonstrate that script is an extremely important factor in predicting transfer success. Specifically, the pre-trained coverage of closely-related languages improves transfer, but only if the target language is written in the same script as its pre-trained relative.

%However, \citet{conneau-etal-2020-emerging} find that embedding spaces learned by pre-trained monolingual models in various languages (and various scripts) share similar structure, suggesting that new vocabulary spaces can be mapped onto a pre-existing encoder.

One adaptation technique is to initialize new subword embeddings that cover the target language, e.g.~by expanding the existing vocabulary with new tokens as necessary, then training the new (randomly initialized) embeddings \cite{chau-etal-2020-parsing, wang-etal-2020-extending}. When transferring a monolingual model to a new language, \citet{artetxe-etal-2020-cross} and \citet{de-vries-nissim-2021-good} instead completely re-initialize the embedding matrix, corresponding to a new subword vocabulary. These embeddings are then trained into alignment with the pre-trained, frozen transformer encoder. We show that this technique is not successful when adapting a multilingual model (Section~\ref{sec:results}).

Other work reuses information in pre-trained embeddings rather than initializing new ones at random. This may include scaling up smaller embedding spaces from models trained on the target language \cite{de-vries-nissim-2021-good, ostendorff_efficient_2023} or copying embeddings from the original vocabulary where there is exact vocabulary overlap \cite{pfeiffer-etal-2021-unks}. When transferring to a target language written in a poorly-covered script, \citet{muller-etal-2021-unseen} show that transliterating the target to the script of a well-covered relative can lead to significant performance gains.

Finally, recent work has proposed more complex methods for mapping source embeddings onto semantically similar ones in the target space either through cross-lingually aligned static word embeddings \cite[e.g. the WESCHEL method;][]{minixhofer-etal-2022-wechsel} or with bilingual lexicons \cite{zeng_greenplm_2023}. In concurrent work to ours, \citet{dobler-de-melo-2023-focus} extend WECHSEL with the \textsc{Focus} method to specialize multilingual vocabularies to a single language. \citet{ostendorff_efficient_2023} use a cross-lingual progressive transfer learning approach to combine information from the source embeddings and a smaller target language model to initialize higher-dimension target embeddings. Unlike earlier initialization methods and our proposed setup, these methods all require additional information outside the source model and often require significant additional compute. We compare one method from this family (\textsc{Focus}) to our proposed heuristic-based initialization schemes.

\section{Vocabulary Replacement \& Embedding Re-initialization}
\label{sec:reinitialization}

Research transferring monolingual models from one language to another (e.g. \citealp{artetxe-etal-2020-cross, de-vries-nissim-2021-good}), has shown that random re-initialization of embeddings +\textsc{Lapt} is sufficient. However, our experiments show that this technique performs poorly when transferring from a multilingual model (Section~\ref{sec:results}). For this reason, we propose several simple techniques for initializing new embeddings based on a qualitative exploration of the embedding space for XLM-R (Section~\ref{sec:embedding_exploration}), and include the more complex \textsc{Focus} technique, developed concurrently with our work, for comparison \citep{dobler-de-melo-2023-focus}.

\subsection{XLM-R Embedding-Space Analysis}
\label{sec:embedding_exploration}

\begin{figure*}[ht]
\centering
\begin{subfigure}{0.32\textwidth}
    \includegraphics[width=\textwidth]{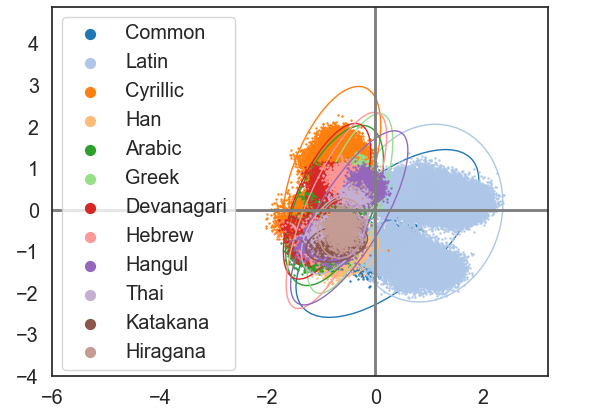}
    \caption{}
    \label{fig:base_xlmr_pca}
\end{subfigure}
\hfill
\begin{subfigure}{0.32\textwidth}
    \includegraphics[width=\textwidth]{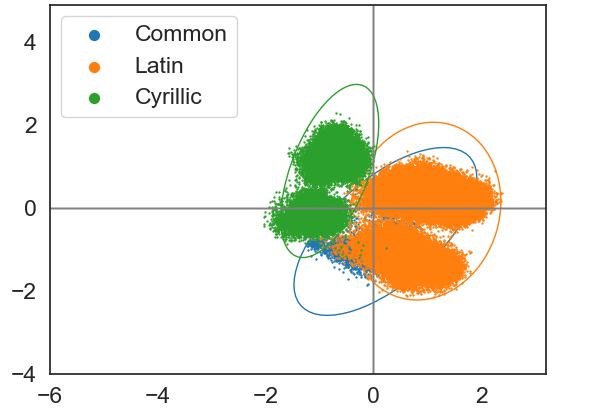}
    \caption{}
    \label{fig:base_uralic_pca}
\end{subfigure}
\hfill
\begin{subfigure}{0.32\textwidth}
    \includegraphics[width=\textwidth]{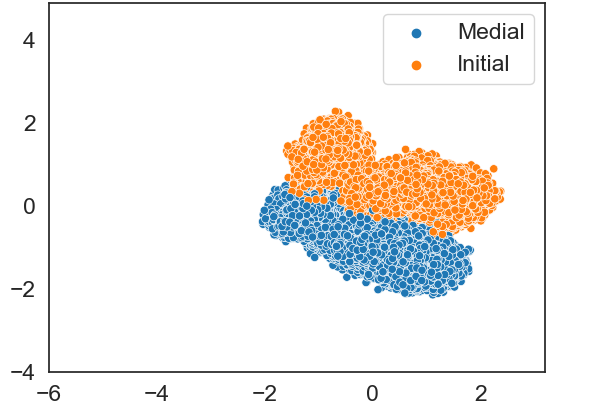}
    \caption{}
    \label{fig:base_position_pca}
\end{subfigure}
\caption{PCA visualizations of the embedding space for XLM-R. Subplots: (a) Distribution of embeddings for the 12 most common Unicode scripts. (b) Plot reduced to only Common, Latin, and Cyrillic scripts for simplicity. (c) Embeddings colored by whether the token begins a word (initial) or occurs in the middle of one (medial)}
\end{figure*}

To better understand the task of initializing new embeddings for a multilingual model, we explore the token-embedding space of XLM-R through PCA projection. Our hypothesis is that multilingual models do not process all languages homogeneously. This seems to be demonstrated in Figures~\ref{fig:base_xlmr_pca} and \ref{fig:base_uralic_pca}, where word embeddings are colored by their respective Unicode script block. We see that the highest-resource scripts in XLM-R (Common, Latin, and Cyrillic) have relatively divergent distributions, while others cluster closer together. This heterogeneity may help explain the finding from \citet{muller-etal-2021-unseen} that pre-trained models do not transfer well to even closely-related target languages if the target script does not match that of the pre-trained relative.

Secondly, each script can be further divided into two sub-distributions, roughly corresponding to a shift in the second principal component. Figure~\ref{fig:base_position_pca} shows that this division corresponds to whether a token is word-initial or word-medial. To preserve whitespace information, SentencePiece tokens include a leading underscore to indicate tokens that should be preceded by a space (word-initial tokens).\footnote{E.g., ``\_the'' and ``the'' are word-initial and word-medial tokens of the same character sequence.} Although the model does not have access to the internal makeup of its tokens, we hypothesize that it learns to discern which tokens can begin a word and which cannot.

Thus when proposing methods to initialize new embeddings for XLM-R, we hypothesize that initializing according script- and position-wise sub-distributions will help to align new vocabulary items with the pre-trained embedding distribution.

\subsection{Embedding Re-initialization Techniques}
\label{sec:reinit_techniques}

\begin{figure*}[ht]
\centering
\begin{subfigure}{0.24\textwidth}
    \includegraphics[width=\textwidth]{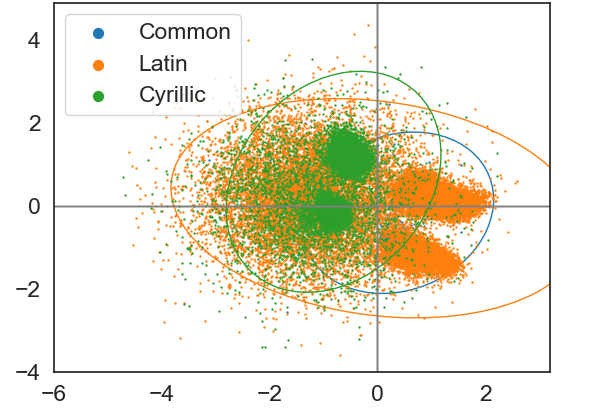}
    \caption{\textsc{ident}}
    \label{fig:uralic_pca_ident}
\end{subfigure}
\hfill
\begin{subfigure}{0.24\textwidth}
    \includegraphics[width=\textwidth]{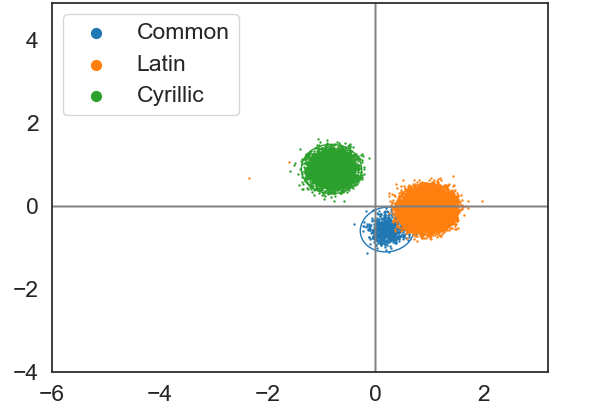}
    \caption{\textsc{script}}
    \label{fig:uralic_pca_script}
\end{subfigure}
\hfill
\begin{subfigure}{0.24\textwidth}
    \includegraphics[width=\textwidth]{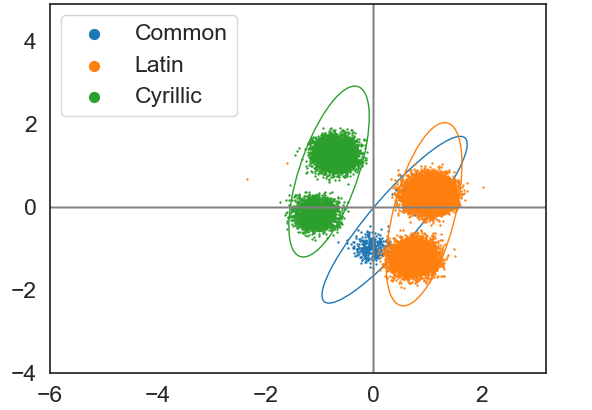}
    \caption{\textsc{script+posn}}
    \label{fig:uralic_pca_script+pos}
\end{subfigure}
\hfill
\begin{subfigure}{0.24\textwidth}
    \includegraphics[width=\textwidth]{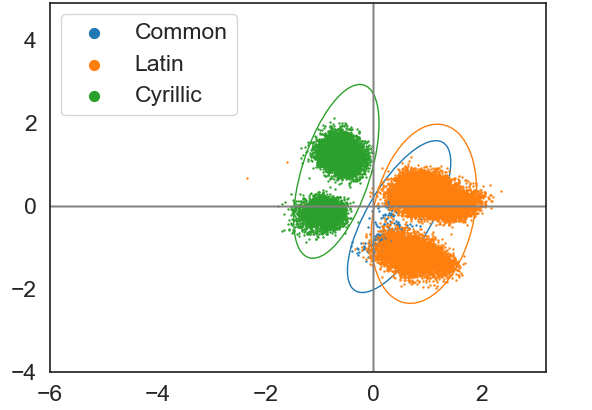}
    \caption{\textsc{script+posn+ident}}
    \label{fig:uralic_pca_script+pos+ident}
\end{subfigure}
\caption{PCA visualizations embedding re-initialized using the heuristic techniques introduced in Section~\ref{sec:reinit_techniques}}
\label{fig:reinit_methods}
\end{figure*}

We now formalize simple techniques for embedding re-initialization based on our exploration of XLM-R's embedding space, as well as one recently proposed technique based on an auxiliary embedding model (\textsc{Focus}). Figure~\ref{fig:reinit_methods} provides PCA visualizations of the re-initialized embeddings from each technique on a subword vocabulary specialized for languages of the Uralic family (we experiment with these languages in Section~\ref{sec:experiments}). The visualization for these languages' respective scripts (Common, Latin, Cyrillic) in the base model can be found in Figure~\ref{fig:base_uralic_pca} for comparison.

\paragraph{Re-initialization by Identity}
\textsc{Reinit-ident} first identifies tokens in the new vocabulary that exactly match a token in the original vocabulary, then sets the new embeddings of shared tokens to be identical to those in the original embedding table (Figure~\ref{fig:uralic_pca_ident}). This is a common approach to preserve information from the original model, even when the other embeddings are randomly re-initialized (e.g., \citealp{pfeiffer-etal-2021-unks}). When identity re-initialization is applied in conjunction with another technique (such as \textsc{Reinit-script}), identity takes precedence.

\paragraph{Re-initialization by Script}
For \textsc{Reinit-script}, all base XLM-R tokens are first categorized by Unicode block, as a stand-in for identifying the script/orthography. We then calculate the mean and standard deviation for each script in the original embedding space. Finally, new token embeddings for each script are distributed according to a Normal distribution with the corresponding mean and standard deviation (Figure~\ref{fig:uralic_pca_script}).

\paragraph{Re-initialization by Position}
\textsc{Reinit-posn} is based on the observation that within each script, embeddings seem to cluster according their word-initial vs. word-medial status (Figure~\ref{fig:base_position_pca}). Similarly to \textsc{Reinit-script}, we identify the mean and standard deviation of embeddings that belong to each category. Because positional status seems to be a sub-cluster within script clusters, we only use \textsc{Reinit-posn} in combination with \textsc{Reinit-script}. The mean and standard deviation for each (script, position) combination is calculated and new embeddings are initialized accordingly (Figure~\ref{fig:uralic_pca_script+pos}).

\paragraph{\textsc{Focus} Re-initialization}
%In addition to the heuristic-based methods we have introduced to initialize new embeddings by categorical aspects like script, we investigate a pre-existing method for embedding transfer and specialization, termed \textsc{Focus} \citep{dobler-de-melo-2023-focus}. 
In addition to the heuristic-based methods introduced above, we investigate a pre-existing method for embedding transfer, termed \textsc{Focus} \citep{dobler-de-melo-2023-focus}. \textsc{Focus} works by extrapolating from the embedding space of an existing model, like our heuristic methods, but further introduces an auxiliary embedding model trained on the new language(s). This auxiliary model (based on FastText; \citealp{bojanowski-etal-2017-enriching}) is used to obtain similarity measures between the new vocabulary items. Embeddings corresponding to overlapping tokens in the new vocabulary keep their values from the source model (\textsc{Reinit-ident}). Completely new tokens are initialized as a weighted combination of the overlapping items, with weights obtained according to similarity in the auxiliary model.

\begin{figure}[ht]
    \begin{center}
    \includegraphics[width=0.28\textwidth]{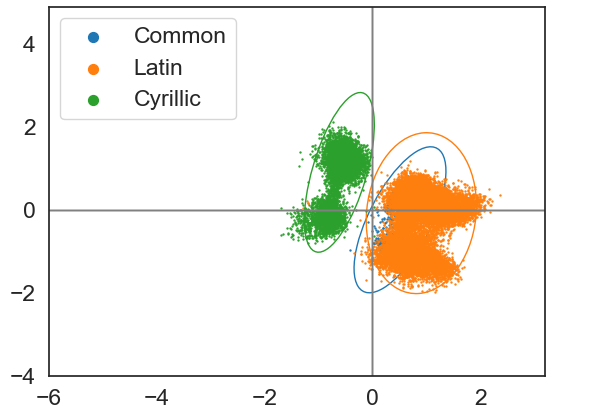}
    \caption{PCA: \textsc{Reinit-focus} embeddings}
    \label{fig:uralic_pca_focus}
    \end{center}
\end{figure}

\paragraph{Random Re-initialization}
Embeddings not initialized through the above methods are initialized according to a Standard Normal Distribution about the origin. This includes the non-overlapping tokens when \textsc{Reinit-ident} is applied on its own, and \textsc{Reinit-random}, where all embeddings are initialized this way.

\paragraph{Inspection of re-initialized embeddings}
Figures~\ref{fig:reinit_methods} and \ref{fig:uralic_pca_focus} show PCA visualizations for the re-initialization techniques described here. Figure~\ref{fig:uralic_pca_ident} shows that while \textsc{Reinit-ident} captures some of the pre-trained embedding structure, a large number also remain randomly scattered throughout the space. \textsc{Reinit-script} (\ref{fig:uralic_pca_script}) initializes all embeddings in a Normal distribution about the centroid for each script, but misses key embedding structure, such as the fact that each script has two position-wise sub-distributions. \textsc{Reinit-script+posn} (\ref{fig:uralic_pca_script+pos}) takes these sub-distributions into account, forming six Normal clusters instead of three.\footnote{Figure~\ref{fig:uralic_position_pca_script+pos} in the Appendix verifies that these clusters capture the initial vs. medial token distinction} Finally, \textsc{Reinit-script+posn+ident} (\ref{fig:uralic_pca_script+pos+ident}) and \textsc{Focus} (\ref{fig:uralic_pca_focus}) give the closest emulation of the original XLM-R embedding structure (\ref{fig:base_uralic_pca}).

\section{Experiments}
\label{sec:experiments}

In our experiments, we replace the large cross-lingual embedding matrix of XLM-R and re-initialize it for a new, language-specific vocabulary. We then conduct \textsc{Lapt} to specialize the model for the new language(s), and evaluate performance on downstream tasks. We consider both multilingual$\rightarrow$monolingual and multilingual$\rightarrow$multilingual transfer scenarios, the latter being transfer to a much smaller set of languages than the original cross-lingual training set. We compare our vocabulary-replacement techniques against the baseline performance of XLM-R off-the-shelf, as well as \textsc{Lapt} while retaining the original, full-sized vocabulary.

Another manipulation we consider is whether the transformer-specific parameters are frozen during \textsc{Lapt}. This follows from the literature on transferring monolingual models, which proposes freezing the encoder parameters and only training the new embedding matrix to mitigate catastrophic forgetting during transfer learning \citep{artetxe-etal-2020-cross, de-vries-nissim-2021-good}. In our tables, we denote \textsc{Lapt} with trainable transformer layers as \textsc{Lapt-full}, and training with the transformer frozen (but trainable embeddings) as \textsc{Lapt-emb}.

\paragraph{Target Languages}
We select our target languages for a wide selection of language families, scripts, typological characteristics, and resource availability, while still having standard evaluation sets for comparison. Training data for all languages is obtained from OSCAR v.22.01 \citep{abadji-etal-2022-towards}. For our lowest-resource languages, supplemental data is obtained from monolingual splits of the OPUS translation corpus \citep{tiedemann-nygaard-2004-opus} and the Johns Hopkins University Bible Corpus \citep{mccarthy-etal-2020-johns}. More data curation details may be found in Appendix~\ref{app:data}.

Our multilingual$\rightarrow$monolingual transfer languages can be found in Table~\ref{mono_pos_table}. In these experiments, the replacement vocabulary and \textsc{Lapt} training are constrained to a single target language. In addition, we include two multilingual$\rightarrow$multilingual experiments. In the first, we simply transfer to the set of languages used in our monolingual experiments. Most of these languages are unrelated and cover a variety of scripts and levels of resource-availability. In the second, we transfer to a set of languages belonging to a single language family --- Uralic. These languages come from the same ancestor language, and share broad grammatical features, but also use both Cyrillic and Latin scripts. These differing settings are designed to demonstrate whether language relatedness has an effect on the success of multilingual vocabulary-replacement techniques.

\paragraph{Vocabulary Replacement / Re-initialization}
When replacing model vocabulary, we train new Sentencepiece models on a subset of the training data. For targets with less than 1GB of data, we use the entire dataset. For those with more, we use a random subset of about 250MB. For multilingual models, we sample 5 million lines according to the same distribution as the training data. All new Sentencepiece models have a total vocabulary size of 32,770 including special tokens. We then initialize the embedding matrix for each new vocabulary according to one or a combination of the techniques described in Section~\ref{sec:reinitialization}.\footnote{The auxiliary FastText model for \textsc{Focus} initialization is trained on the same set as the vocabulary}

\paragraph{Training}
All of our experiments use XLM-R as a starting point (base size; \citealp{conneau-etal-2020-unsupervised}). We conduct \textsc{Lapt} for 100k training steps, with evaluation checkpoints every 1000 steps. For \textsc{Lapt-full} experiments, the transformer blocks are frozen for the first 10k steps, then unfrozen for the last 90k, so that the model does not overfit to initial (possibly poor) embedding initializations. For \textsc{Lapt-emb} experiments, transformer blocks remain frozen throughout training. The checkpoint obtaining the best MLM loss on a development set is selected for task fine-tuning and evaluation.

For multilingual training, we sample languages according to a multinomial distribution parameterized by $\alpha=0.2$, following \citet{conneau-lample-2019-cross}, \citet{conneau-etal-2020-unsupervised}, i.a. Languages are sampled sentence-wise rather than batch-wise.

\paragraph{Evaluation}
We evaluate model quality with POS-tagging and NER tasks. For each task and each language, the trained model is fine-tuned on task training data until evaluation set convergence or the maximum number of epochs is reached, across four random seeds. POS performance is evaluated on  Universal Dependencies (UD) treebanks \cite{de-marneffe-etal-2021-universal}, and NER is measured on the WikiAnn benchmark \cite{pan-etal-2017-cross}.

\section{Results}
\label{sec:results}

\begin{table*}[ht]
    \centering
    \resizebox{0.8\textwidth}{!}{
    \begin{tabular}{llccccccccc}
        \toprule
        \textsc{Lapt} & \textsc{Reinit} & Armenian & Basque & Erzya & Estonian & Hebrew & Russian & North Sami & Telugu & Avg \\
        \midrule
        * & * & 93.4 $\pm$ 2.2 & 95.1 $\pm$ 0.7 & 56.3 $\pm$ 5.3 & \underline{95.6 $\pm$ 0.1} & \underline{97.5 $\pm$ 0.1} & \underline{98.6 $\pm$ 0.1} & 71.2 $\pm$ 1.8 & 83.8 $\pm$ 0.1 & 86.4 \\
        \textsc{full} & * & - & - & \underline{85.1 $\pm$ 1.8} & - & 97.5 $\pm$ 0.1 & - & - & \underline{91.4 $\pm$ 4.3} & - \\
        \midrule
        \textsc{full} & \textsc{focus+ident} & \textbf{92.3 $\pm$ 1.9} & \underline{\textbf{96.0 $\pm$ 0.6}} & 76.1 $\pm$ 2.0 & \textbf{95.1 $\pm$ 0.3} & \textbf{97.2 $\pm$ 0.1} & \textbf{98.4 $\pm$ 0.1} & \underline{\textbf{92.1 $\pm$ 0.8}} & \textbf{86.9 $\pm$ 3.5} & \underline{\textbf{91.7}} \\
        \textsc{full} & \textsc{script+posn+ident} & \underline{\textbf{93.1 $\pm$ 1.7}} & 93.8 $\pm$ 0.5 & \textbf{79.0 $\pm$ 0.7} & 94.0 $\pm$ 0.2 & 96.7 $\pm$ 0.1 & 98.2 $\pm$ 0.04 & 86.9 $\pm$ 0.7 & \textbf{88.5 $\pm$ 3.2} & 91.3 \\
        \textsc{full} & \textsc{script+ident} & 91.7 $\pm$ 1.9 & 93.6 $\pm$ 0.3 & 70.8 $\pm$ 12.8 & 94.0 $\pm$ 0.1 & 96.7 $\pm$ 0.1 & 98.1 $\pm$ 0.1 & 83.4 $\pm$ 1.3 & \textbf{87.1 $\pm$ 3.4} & 89.4 \\
        \textsc{full} & \textsc{script+posn} & 90.9 $\pm$ 2.0 & 92.1 $\pm$ 0.7 & 74.6 $\pm$ 2.2 & 90.4 $\pm$ 0.6 & 95.4 $\pm$ 0.1 & 97.2 $\pm$ 0.02 & 78.7 $\pm$ 0.5 & \textbf{87.5 $\pm$ 1.4} & 88.3 \\
        \textsc{full} & \textsc{script} & 89.6 $\pm$ 1.5 & 90.9 $\pm$ 0.2 & 71.5 $\pm$ 2.1 & 89.4 $\pm$ 0.9 & 95.0 $\pm$ 0.05 & 96.9 $\pm$ 0.03 & 77.9 $\pm$ 0.2 & 84.0 $\pm$ 1.5 & 86.9 \\
        \textsc{full} & \textsc{ident} & 81.6 $\pm$ 0.4 & 83.6 $\pm$ 0.6 & 59.1 $\pm$ 3.1 & 86.4 $\pm$ 0.4 & 91.1 $\pm$ 0.1 & 96.2 $\pm$ 0.04 & 70.7 $\pm$ 0.5 & 78.0 $\pm$ 2.5 & 80.9 \\
        \textsc{full} & \textsc{random} & 67.4 $\pm$ 2.0 & 72.7 $\pm$ 0.6 & 53.3 $\pm$ 2.8 & 72.0 $\pm$ 0.1 & 81.0 $\pm$ 0.6 & 86.5 $\pm$ 0.6 & 64.7 $\pm$ 0.9 & 76.4 $\pm$ 1.0 & 72.4 \\
        \midrule
        \textsc{emb} & \textsc{focus+ident} & \textbf{92.3 $\pm$ 1.7} & \textbf{95.1 $\pm$ 0.6} & 48.6 $\pm$ 0.1 & \textbf{94.5 $\pm$ 0.05} & \textbf{96.9 $\pm$ 0.3} & \textbf{98.3 $\pm$ 0.04} & \textbf{73.6 $\pm$ 1.6} & \textbf{86.2 $\pm$ 3.8} & \textbf{84.8} \\
        \textsc{emb} & \textsc{script+posn+ident} & 87.6 $\pm$ 1.3 & 88.2 $\pm$ 0.7 & \textbf{55.6 $\pm$ 4.8} & 89.6 $\pm$ 0.1 & 95.3 $\pm$ 0.1 & 97.1 $\pm$ 0.05 & 69.8 $\pm$ 1.4 & 81.8 $\pm$ 1.2 & 82.5 \\
        \textsc{emb} & \textsc{script+ident} & 87.7 $\pm$ 1.8 & 87.9 $\pm$ 0.4 & \textbf{53.8 $\pm$ 5.4} & 89.2 $\pm$ 0.5 & 95.2 $\pm$ 0.1 & 97.0 $\pm$ 0.1 & 68.6 $\pm$ 1.8 & 82.0 $\pm$ 1.3 & 82.0 \\
        \textsc{emb} & \textsc{script+posn} & 56.5 $\pm$ 7.6 & 61.3 $\pm$ 12.0 & 48.7 $\pm$ 0.1 & 71.4 $\pm$ 1.4 & 82.5 $\pm$ 0.3 & 92.1 $\pm$ 0.4 & 59.8 $\pm$ 1.5 & 70.1 $\pm$ 7.4 & 69.4 \\
        \textsc{emb} & \textsc{script} & 47.6 $\pm$ 6.4 & 59.6 $\pm$ 8.1 & 48.6 $\pm$ 0.1  & 65.7 $\pm$ 5.2 & 80.4 $\pm$ 2.2 & 89.7 $\pm$ 1.0 & 55.5 $\pm$ 5.0 & 73.4 $\pm$ 5.5 & 67.6 \\
        \textsc{emb} & \textsc{ident} & 80.3 $\pm$ 1.1 & 80.1 $\pm$ 0.6 & 47.9 $\pm$ 1.5 & 82.5 $\pm$ 1.8 & 88.7 $\pm$ 0.2 & 95.2 $\pm$ 0.4 & 60.6 $\pm$ 1.2 & 76.6 $\pm$ 1.4 & 75.9 \\
        \textsc{emb} & \textsc{random} & 47.6 $\pm$ 1.8 & 55.2 $\pm$ 2.8 & 46.3 $\pm$ 0.2 & 63.5 $\pm$ 1.8 & 67.6 $\pm$ 2.5 & 80.2 $\pm$ 0.6 & 44.7 $\pm$ 4.0 & 56.7 $\pm$ 6.7 & 59.2 \\
        \bottomrule
    \end{tabular}}
    \caption{Monolingual Language-Adaptive Pre-Training (\textsc{Lapt}): POS tagging accuracy after fine-tuning. * indicates XLM-R off-the-shelf. Within each division, best result and results within 1 standard deviation are bolded; overall best result indicated with added underline. Best result determined by \textit{mean - stdev}. \textsc{Lapt} with full XLM-R vocab only conducted for three languages due to prohibitive computational cost}
    \label{mono_pos_table}
\end{table*}

\begin{table*}[ht]
    \centering
    \resizebox{0.8\textwidth}{!}{
    \begin{tabular}{llccccccccc}
        \toprule
        \textsc{Lapt} & \textsc{Reinit} & Armenian & Basque & Erzya & Estonian & Hebrew & Russian & Telugu & Avg \\
        \midrule
        * & * & 94.1 $\pm$ 0.1 & 94.3 $\pm$ 0.1 & 89.5 $\pm$ 0.6 & \underline{93.3 $\pm$ 0.2} & 85.9 $\pm$ 0.1 & \underline{90.9 $\pm$ 0.2} & 85.4 $\pm$ 0.5 & 90.5 \\
        \textsc{full} & * & - & - & \underline{91.8 $\pm$ 0.5} & - & \underline{86.9 $\pm$ 0.1} & - & 86.6 $\pm$ 1.9 & - \\
        \midrule
        \textsc{full} & \textsc{focus+ident} & \underline{\textbf{95.1 $\pm$ 0.9}} & \underline{\textbf{94.9 $\pm$ 0.4}} & \textbf{89.9 $\pm$ 0.8} & \textbf{92.6 $\pm$ 0.2} & \textbf{86.2 $\pm$ 0.3} & \textbf{90.6 $\pm$ 0.1} & \underline{\textbf{87.7 $\pm$ 0.5}} & \textbf{91.0} \\
        \textsc{full} & \textsc{script+posn+ident} & 93.9 $\pm$ 0.1 & 94.3 $\pm$ 0.2 & \textbf{90.2 $\pm$ 0.7} & 92.0 $\pm$ 0.3 & 83.2 $\pm$ 0.4 & 89.8 $\pm$ 0.2 & 83.5 $\pm$ 1.8 & 89.6 \\
        \textsc{full} & \textsc{script+ident} & 93.8 $\pm$ 0.3 & 94.3 $\pm$ 0.1 & \textbf{89.8 $\pm$ 0.2} & 89.3 $\pm$ 0.2 & 83.4 $\pm$ 0.3 & 89.4 $\pm$ 0.2 & 84.0 $\pm$ 0.5 & 89.5 \\
        \textsc{full} & \textsc{script+posn} & 92.0 $\pm$ 0.6 & 92.1 $\pm$ 0.04 & 89.1 $\pm$ 0.5 & 88.3 $\pm$ 0.4 & 78.7 $\pm$ 0.1 & 86.5 $\pm$ 0.1 & 81.0 $\pm$ 0.9 & 86.8 \\
        \textsc{full} & \textsc{script} & 91.4 $\pm$ 0.4 & 91.1 $\pm$ 0.1 & 87.7 $\pm$ 0.5 & 87.5 $\pm$ 0.2 & 78.5 $\pm$ 0.2 & 85.7 $\pm$ 0.1 & 79.6 $\pm$ 1.1 & 85.9 \\
        \textsc{full} & \textsc{ident} & 86.2 $\pm$ 0.4 & 90.7 $\pm$ 0.2 & 79.0 $\pm$ 0.6 & 89.3 $\pm$ 0.2 & 72.0 $\pm$ 0.4 & 86.7 $\pm$ 0.1 & 69.3 $\pm$ 0.4 & 81.9 \\
        \textsc{full} & \textsc{random} & 74.1 $\pm$ 1.4 & 81.5 $\pm$ 0.3 & 72.6 $\pm$ 3.3 & 45.8 $\pm$ 27.2 & 54.4 $\pm$ 0.9 & 70.3 $\pm$ 0.7 & 47.2 $\pm$ 8.2 & 63.7 \\
        \midrule
        \textsc{emb} & \textsc{focus+ident} & \textbf{93.5 $\pm$ 0.5} & \textbf{94.2 $\pm$ 0.2} & 81.7 $\pm$ 2.2 & \textbf{92.0 $\pm$ 0.2} & \textbf{84.9 $\pm$ 0.1} & \textbf{90.3 $\pm$ 0.1} & \textbf{86.1 $\pm$ 0.3} & \textbf{89.0} \\
        \textsc{emb} & \textsc{script+posn+ident} & 91.5 $\pm$ 0.2 & 92.3 $\pm$ 0.1 & \textbf{87.2 $\pm$ 0.3} & 89.8 $\pm$ 0.2 & 79.1 $\pm$ 0.2 & 88.9 $\pm$ 0.1 & 74.1 $\pm$ 1.2 & 86.1 \\
        \textsc{emb} & \textsc{script+ident} & 90.9 $\pm$ 0.3 & 92.0 $\pm$ 0.3 & 86.1 $\pm$ 1.0 & 89.6 $\pm$ 0.3 & 78.7 $\pm$ 0.3 & 88.6 $\pm$ 0.1 & 79.1 $\pm$ 0.5 & 86.4 \\
        \textsc{emb} & \textsc{script+posn} & 86.5 $\pm$ 0.4 & 87.3 $\pm$ 0.3 & 84.1 $\pm$ 1.2 & 81.8 $\pm$ 0.8 & 71.0 $\pm$ 0.9 & 81.0 $\pm$ 0.2 & 64.3 $\pm$ 1.9 & 79.4 \\
        \textsc{emb} & \textsc{script} & 83.9 $\pm$ 0.4 & 73.0 $\pm$ 0.8 & 84.0 $\pm$ 1.2 & 79.5 $\pm$ 0.9 & 67.8 $\pm$ 0.6 & 77.4 $\pm$ 0.2 & 56.8 $\pm$ 3.2 & 74.6 \\
        \textsc{emb} & \textsc{ident} & 80.9 $\pm$ 0.8 & 87.9 $\pm$ 0.4 & 61.8 $\pm$ 3.8 & 85.3 $\pm$ 0.3 & 64.8 $\pm$ 1.4 & 84.8 $\pm$ 0.4 & 54.9 $\pm$ 1.5 & 74.3 \\
        \textsc{emb} & \textsc{random} & 59.6 $\pm$ 2.5 & 0.0 $\pm$ 0.0 & 51.8 $\pm$ 2.7 & 0.0 $\pm$ 0.0 & 17.1 $\pm$ 17.2 & 47.5 $\pm$ 6.9 & 22.4 $\pm$ 5.5 & 28.3 \\
        \bottomrule
    \end{tabular}}
    \caption{Monolingual \textsc{Lapt}: entity-wise NER F1 score after fine-tuning. A score of 0.0 results from the model learning to output only class \texttt{O} (not a named entity) which is the majority class. Sami does not have enough NER data for fine-tuning}
    \label{mono_ner_table}
\end{table*}

\begin{table*}[ht]
    \centering
    \resizebox{0.8\textwidth}{!}{
    \begin{tabular}{llccccccccc}
        \toprule
        \textsc{Lapt} & \textsc{Reinit} & Armenian & Basque & Erzya & Estonian & Hebrew & Russian & North Sami & Telugu & Avg \\
        \midrule
        * & * & 93.4 $\pm$ 2.2 & 95.1 $\pm$ 0.7 & 56.3 $\pm$ 5.3 & \underline{95.6 $\pm$ 0.1} & \underline{97.5 $\pm$ 0.1} & 98.6 $\pm$ 0.1 & 71.2 $\pm$ 1.8 & 83.8 $\pm$ 0.1 & 86.4 \\
        \textsc{full} & * & 91.3 $\pm$ 0.1 & \underline{95.9 $\pm$ 0.6} & 71.7 $\pm$ 5.3 & 95.5 $\pm$ 0.2 & 97.4 $\pm$ 0.2 & \underline{98.6 $\pm$ 0.04} & \underline{80.6 $\pm$ 1.4} & 89.7 $\pm$ 3.6 & \underline{90.1} \\
        \midrule
        \textsc{full} & \textsc{focus+ident} & 91.0 $\pm$ 0.1 & \textbf{95.8 $\pm$ 0.1} & \underline{\textbf{72.5 $\pm$ 1.3}} & \textbf{95.5 $\pm$ 0.2} & \textbf{97.1 $\pm$ 0.1} & \textbf{98.4 $\pm$ 0.03} & \textbf{80.4 $\pm$ 1.2} & \textbf{89.4 $\pm$ 3.2} & \textbf{90.0} \\
        \textsc{full} & \textsc{script+posn+ident} & \textbf{92.9 $\pm$ 2.1} & 95.0 $\pm$ 0.6 & \textbf{63.6 $\pm$ 9.8} & 94.8 $\pm$ 0.3 & 97.0 $\pm$ 0.1 & \textbf{98.4 $\pm$ 0.04} & \textbf{80.4 $\pm$ 1.1} & \textbf{89.6 $\pm$ 2.6} & 89.0 \\
        \textsc{full} & \textsc{script+ident} & \textbf{93.8 $\pm$ 1.8} & 95.3 $\pm$ 0.03 & \textbf{66.1 $\pm$ 10.2} & 94.7 $\pm$ 0.2 & \textbf{97.1 $\pm$ 0.1} & \textbf{98.4 $\pm$ 0.03} & \textbf{80.1 $\pm$ 1.2} & \underline{\textbf{91.7 $\pm$ 0.8}} & 89.7 \\
        \textsc{full} & \textsc{script+posn} & 85.3 $\pm$ 3.5 & 87.9 $\pm$ 3.5 & 70.5 $\pm$ 1.5 & 89.0 $\pm$ 0.8 & 93.7 $\pm$ 0.6 & 97.2 $\pm$ 0.01 & 72.8 $\pm$ 2.1 & 81.6 $\pm$ 0.4 & 84.7 \\
        \textsc{full} & \textsc{script} & 83.3 $\pm$ 1.9 & 85.8 $\pm$ 2.7 & 66.6 $\pm$ 1.9 & 85.4 $\pm$ 1.7 & 90.5 $\pm$ 0.8 & 96.8 $\pm$ 0.03 & 68.6 $\pm$ 1.1 & 81.0 $\pm$ 0.3 & 82.2 \\
        \textsc{full} & \textsc{ident} & \underline{\textbf{93.2 $\pm$ 0.7}} & 93.0 $\pm$ 0.5 & 58.1 $\pm$ 0.9 & 93.6 $\pm$ 0.2 & 96.6 $\pm$ 0.1 & 98.3 $\pm$ 0.03 & 71.5 $\pm$ 1.2 & 89.0 $\pm$ 4.1 & 86.7 \\
        \textsc{full} & \textsc{random} & 64.5 $\pm$ 2.9 & 67.4 $\pm$ 0.4 & 50.0 $\pm$ 4.6 & 71.9 $\pm$ 0.3 & 80.0 $\pm$ 0.8 & 84.6 $\pm$ 0.9 & 62.7 $\pm$ 0.5 & 75.0 $\pm$ 6.2 & 70.2 \\
        \midrule
        \textsc{emb} & \textsc{focus+ident} & \textbf{93.1 $\pm$ 2.2} & \textbf{95.2 $\pm$ 0.7} & \textbf{63.7 $\pm$ 2.0} & \textbf{94.7 $\pm$ 0.1} & \textbf{97.1 $\pm$ 0.04} & \textbf{98.5 $\pm$ 0.03} & 71.2 $\pm$ 2.1 & \textbf{87.5 $\pm$ 2.9} & \textbf{86.8} \\
        \textsc{emb} & \textsc{script+posn+ident} & \textbf{91.3 $\pm$ 1.6} & 93.5 $\pm$ 0.6 & 57.2 $\pm$ 7.0 & 93.5 $\pm$ 0.1 & 96.7 $\pm$ 0.03 & 98.3 $\pm$ 0.1 & \textbf{74.5 $\pm$ 1.1} & \textbf{85.6 $\pm$ 2.9} & 85.6 \\
        \textsc{emb} & \textsc{script+ident} & \textbf{92.2 $\pm$ 2.0} & 93.2 $\pm$ 0.7 & 58.5 $\pm$ 6.9 & 93.3 $\pm$ 0.1 & 96.9 $\pm$ 0.1 & 98.3 $\pm$ 0.02 & \textbf{72.0 $\pm$ 3.0} & \textbf{86.5 $\pm$ 2.4} & 85.5 \\
        \textsc{emb} & \textsc{script+posn} & 61.5 $\pm$ 1.9 & 76.0 $\pm$ 1.3 & 51.9 $\pm$ 3.1 & 75.7 $\pm$ 0.2 & 87.2 $\pm$ 1.2 & 95.3 $\pm$ 0.3 & 65.3 $\pm$ 0.2 & 77.3 $\pm$ 0.3 & 75.5 \\
        \textsc{emb} & \textsc{script} & 44.7 $\pm$ 0.0 & 71.0 $\pm$ 1.0 & 48.5 $\pm$ 0.2 & 73.5 $\pm$ 2.2 & 83.6 $\pm$ 0.3 & 93.5 $\pm$ 0.5 & 63.8 $\pm$ 1.4 & 77.7 $\pm$ 0.5 & 73.1 \\
        \textsc{emb} & \textsc{ident} & 89.4 $\pm$ 0.8 & 90.5 $\pm$ 0.6 & 49.3 $\pm$ 4.6 & 91.8 $\pm$ 0.5 & 96.2 $\pm$ 0.1 & 98.1 $\pm$ 0.1 & 65.6 $\pm$ 1.1 & 84.0 $\pm$ 1.7 & 82.2 \\
        \textsc{emb} & \textsc{random} & 48.7 $\pm$ 2.4 & 61.2 $\pm$ 5.6 & 46.0 $\pm$ 0.3 & 66.3 $\pm$ 3.9 & 73.7 $\pm$ 3.4 & 85.1 $\pm$ 1.2 & 44.7 $\pm$ 4.6 & 67.5 $\pm$ 5.0 & 63.5 \\
        \bottomrule
    \end{tabular}}
    \caption{Multilingual \textsc{Lapt}: POS tagging accuracy after fine-tuning}
    \label{multi_pos_table}
\end{table*}

\begin{table*}[ht]
    \centering
    \resizebox{0.8\textwidth}{!}{
    \begin{tabular}{llccccccccc}
        \toprule
        \textsc{Lapt} & \textsc{Reinit} & Armenian & Basque & Erzya & Estonian & Hebrew & Russian & Telugu & Avg \\
        \midrule
        * & * & 94.1 $\pm$ 0.1 & 94.3 $\pm$ 0.1 & 89.5 $\pm$ 0.6 & 93.3 $\pm$ 0.2 & 85.9 $\pm$ 0.1 & 90.9 $\pm$ 0.2 & 85.4 $\pm$ 0.5 & 90.5 \\
        \textsc{full} & * & 94.0 $\pm$ 0.5 & \underline{94.5 $\pm$ 0.2} & \underline{90.5 $\pm$ 0.3} & \underline{93.7 $\pm$ 0.2} & \underline{86.2 $\pm$ 0.1} & \underline{91.1 $\pm$ 0.2} & \underline{85.9 $\pm$ 0.7} & \underline{90.9} \\
        \midrule
        \textsc{full} & \textsc{focus+ident} & \textbf{94.2 $\pm$ 0.3} & \textbf{94.0 $\pm$ 0.2} & \textbf{89.6 $\pm$ 1.0} & \textbf{92.0 $\pm$ 0.5} & \textbf{85.2 $\pm$ 0.1} & \textbf{90.0 $\pm$ 0.5} & \textbf{85.4 $\pm$ 0.4} & \textbf{90.1} \\
        \textsc{full} & \textsc{script+posn+ident} & \textbf{94.1 $\pm$ 0.2} & \textbf{94.0 $\pm$ 0.1} & 88.8 $\pm$ 0.9 & \textbf{92.3 $\pm$ 0.1} & 85.0 $\pm$ 0.2 & \textbf{90.4 $\pm$ 0.1} & 84.8 $\pm$ 0.4 & 89.9 \\
        \textsc{full} & \textsc{script+ident} & \underline{\textbf{94.2 $\pm$ 0.2}} & \textbf{94.1 $\pm$ 0.2} & \textbf{90.1 $\pm$ 0.6} & \textbf{92.4 $\pm$ 0.1} & 84.9 $\pm$ 0.3 & 90.3 $\pm$ 0.1 & 84.5 $\pm$ 0.2 & 90.0 \\
        \textsc{full} & \textsc{script+posn} & 91.2 $\pm$ 0.5 & 91.5 $\pm$ 0.1 & 88.9 $\pm$ 0.5 & 88.4 $\pm$ 0.4 & 77.3 $\pm$ 0.4 & 86.3 $\pm$ 0.1 & 76.2 $\pm$ 0.4 & 85.7 \\
        \textsc{full} & \textsc{script} & 90.9 $\pm$ 0.1 & 91.3 $\pm$ 0.3 & 86.4 $\pm$ 1.9 & 87.7 $\pm$ 0.2 & 75.8 $\pm$ 0.3 & 85.7 $\pm$ 0.1 & 75.1 $\pm$ 0.9 & 84.7 \\
        \textsc{full} & \textsc{ident} & 93.2 $\pm$ 0.1 & 93.4 $\pm$ 0.2 & 80.9 $\pm$ 2.4 & 91.5 $\pm$ 0.4 & 83.5 $\pm$ 0.3 & 89.8 $\pm$ 0.1 & 83.2 $\pm$ 0.5 & 87.9 \\
        \textsc{full} & \textsc{random} & 69.9 $\pm$ 4.4 & 80.9 $\pm$ 0.5 & 75.2 $\pm$ 1.5 & 70.5 $\pm$ 2.1 & 37.7 $\pm$ 21.8 & 68.6 $\pm$ 0.7 & 42.1 $\pm$ 1.6 & 63.6 \\
        \midrule
        \textsc{emb} & \textsc{focus+ident} & \textbf{93.9 $\pm$ 0.3} & \textbf{93.7 $\pm$ 0.2} & \textbf{89.7 $\pm$ 0.4} & \textbf{91.9 $\pm$ 0.4} & \textbf{84.8 $\pm$ 0.2} & \textbf{89.9 $\pm$ 0.3} & \textbf{85.2 $\pm$ 0.5} & \textbf{89.9} \\
        \textsc{emb} & \textsc{script+posn+ident} & \textbf{93.7 $\pm$ 0.2} & 93.5 $\pm$ 0.1 & 87.2 $\pm$ 1.0 & \textbf{91.9 $\pm$ 0.2} & 84.0 $\pm$ 0.2 & \textbf{89.9 $\pm$ 0.2} & 84.0 $\pm$ 0.5 & 89.2 \\
        \textsc{emb} & \textsc{script+ident} & 93.3 $\pm$ 0.5 & 93.4 $\pm$ 0.2 & 85.8 $\pm$ 1.4 & \textbf{91.9 $\pm$ 0.3} & 83.7 $\pm$ 0.2 & \textbf{89.9 $\pm$ 0.1} & 82.5 $\pm$ 1.3 & 88.7 \\
        \textsc{emb} & \textsc{script+posn} & 87.5 $\pm$ 0.3 & 88.8 $\pm$ 0.3 & 81.0 $\pm$ 3.1 & 84.8 $\pm$ 0.4 & 72.8 $\pm$ 0.1 & 82.7 $\pm$ 0.3 & 67.1 $\pm$ 1.3 & 80.7 \\
        \textsc{emb} & \textsc{script} & 85.2 $\pm$ 0.3 & 81.3 $\pm$ 7.1 & 80.0 $\pm$ 1.1 & 84.3 $\pm$ 0.3 & 68.3 $\pm$ 0.9 & 80.6 $\pm$ 1.0 & 59.7 $\pm$ 3.5 & 77.1 \\
        \textsc{emb} & \textsc{ident} & 91.2 $\pm$ 0.3 & 92.3 $\pm$ 0.2 & 76.7 $\pm$ 1.3 & 90.8 $\pm$ 0.3 & 81.6 $\pm$ 0.2 & 89.3 $\pm$ 0.2 & 78.6 $\pm$ 1.8 & 85.8 \\
        \textsc{emb} & \textsc{random} & 62.8 $\pm$ 0.9 & 74.9 $\pm$ 1.6 & 66.1 $\pm$ 1.1 & 62.7 $\pm$ 1.9 & 23.9 $\pm$ 18.2 & 53.1 $\pm$ 4.7 & 37.7 $\pm$ 2.6 & 54.4 \\
        \bottomrule
    \end{tabular}}
    \caption{Multilingual \textsc{Lapt}: entity-wise NER F1 score after fine-tuning}
    \label{multi_ner_table}
\end{table*}

The results for monolingual adaptation can be found in Tables~\ref{mono_pos_table}-\ref{mono_ner_table} and general multilingual adaptation in Tables~\ref{multi_pos_table}-\ref{multi_ner_table}. Because the results for multilingual adaptation to the Uralic family mostly echo overall trends, we provide these results in Appendix~\ref{app:uralic_results}.\footnote{While training on related languages may be beneficial for low-resource Uralic languages like Erzya, family-based training vs. general multilingual training does not seem to alter the relative ranking of embedding initialization techniques, which is our primary research interest} In order to adhere to our overall computational budget, we only conduct full-vocabulary \textsc{Lapt} experiments for three languages in the monolingual setting.\footnote{We select Erzya, Telugu, and Hebrew for these full-size experiments, spanning very-low, low, and medium resource-availability levels}

We first note that across re-initialization methods, \textsc{Lapt-full} always outperforms \textsc{Lapt-emb}. I.e.~training with trainable transformer layers outperforms training with frozen ones, despite the risk of catastrophic forgetting with the former. This trend persists across monolingual and multilingual experiments. For example, \textsc{Reinit-focus+ident} shows a 6.9 average POS accuracy drop between \textsc{Lapt-full} and \textsc{Lapt-emb} (Table~\ref{mono_pos_table}).

Second, although \textsc{Focus} is the best performing re-initialization method when averaged across languages, for individual languages, it does not perform significantly differently than script-based methods. For instance, Armenian and Telugu POS tagging with script-based initialization performs on-par with or better than \textsc{Focus} (Tables~\ref{mono_pos_table}, \ref{multi_pos_table}).\footnote{Overall performance/ranking of \textsc{script+posn+ident} vs. \textsc{script+ident} remains uncertain. For \textsc{Lapt-full} averaged across languages, the former performs better in 2/3 POS settings, but only 1/3 NER settings} In the case of the very low-resource language Erzya, script-based methods mostly outperform \textsc{Focus}.\footnote{However, script-based methods show significant variation on Erzya POS after multilingual training (Table~\ref{multi_pos_table})}

Third, for the languages with the largest amount of data in XLM-R (Estonian, Hebrew, and Russian), the off-the-shelf performance of XLM-R (top row) is slightly better than any re-initialization method. This is not unexpected, since we can expect the highest-resource languages in XLM-R to receive adequate vocabulary coverage, and their embeddings are likely the most robustly trained.

\begin{figure}[ht]
    \begin{center}
    \includegraphics[width=0.5\textwidth]{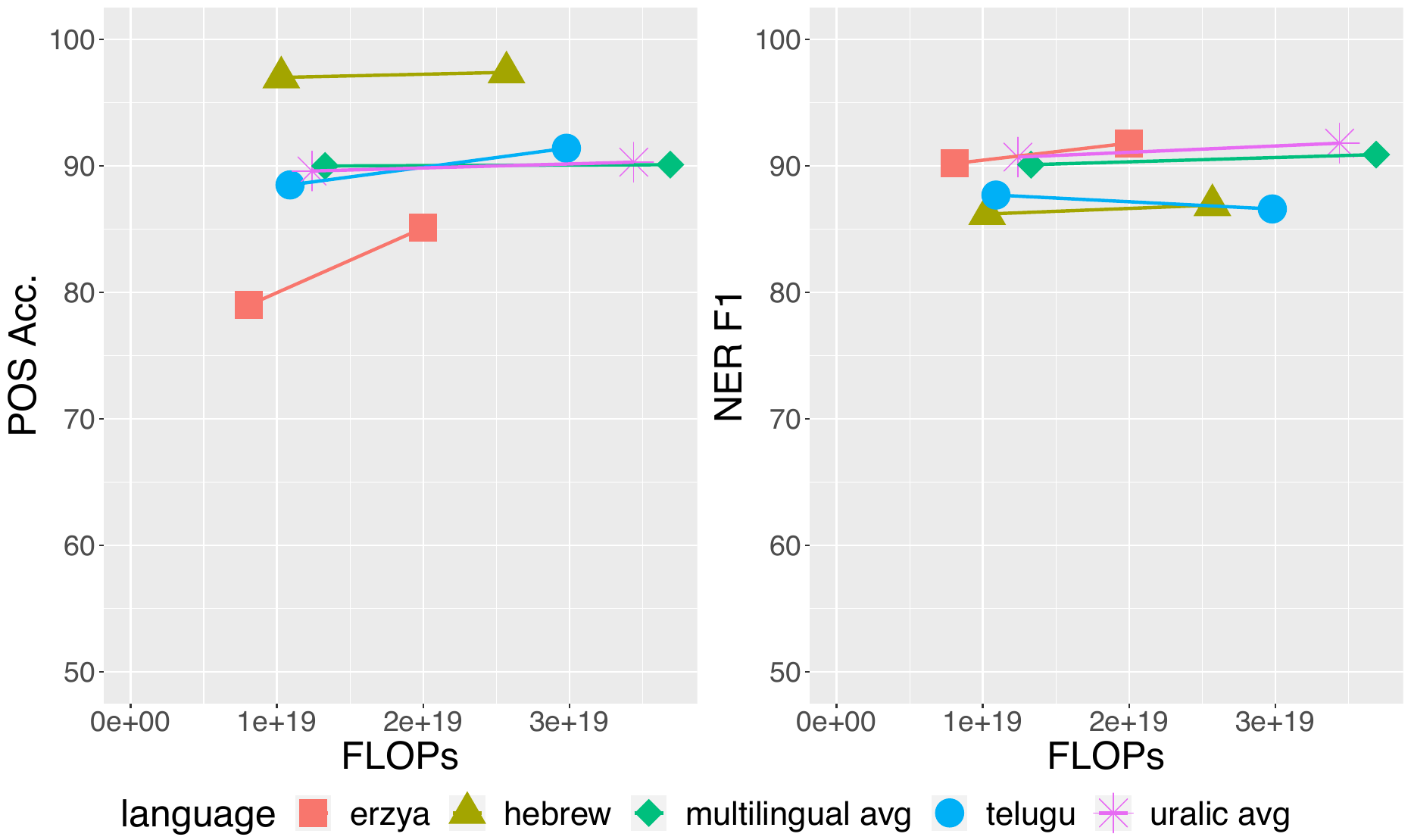}
    \caption{Evaluation scores plotted against total floating point operations of \textsc{Lapt} (computational cost). Left point represents cost of \textsc{Lapt} with reduced vocabulary, right point with full vocabulary}
    \label{fig:score_vs_flops}
    \end{center}
\end{figure}

Finally, \textsc{Lapt} with the full, original XLM-R vocabulary, results in marginally better performance than other techniques. On one hand, this might be surprising given the inefficiency with which cross-lingual vocabularies often tokenize low-resource languages \citep{acs-2019-exploring}. On the other hand, these original pre-trained embeddings are also likely robustly aligned with the transformer encoder, which might contribute to slightly better performance.

Part of the motivation for this work, however, is to investigate \textit{efficient} ways to specialize multilingual models. \textsc{Lapt} with the full XLM-R vocabulary is much more computationally costly than training new vocabulary. Figure~\ref{fig:score_vs_flops} shows the tradeoff between computation (in FLOPs) and performance gain in our experiments: the (often) small gains in performance we see from fine-tuning with the original vocabulary come at the cost of two to three times more FLOPs during adaptation.

Erzya POS performance provides one exception to the pattern of full-vocab \textsc{Lapt} providing only marginal benefits (85.1 accuracy with the full vocabulary vs. 79.0 with the reduced vocabulary). This seems surprising, given Erzya is not included in XLM-R's pre-training data, and intuitively should benefit the most from a specialized vocabulary. It could be that the reduced vocabulary size of 32k is sub-optimal for this particular target language, and/or that the new vocabulary does not overlap enough with the original (full-size) one to inherit useful Cyrillic-script embeddings. Investigating the dynamics of target vocabulary size during vocabulary specialization would be a fruitful direction for future work.

\section{Discussion}
\label{sec:discussion}

\paragraph{Embedding-only training is inadequate for multilingual model transfer}
Our experiments show that language transfer methods developed for monolingual models, which freeze the transformer blocks and re-train only the embedding matrix \citep{artetxe-etal-2020-cross, de-vries-nissim-2021-good}, yield poor results when transferring a multilingual model. This work in the monolingual literature not only keeps transformer layers frozen, but initializes new embeddings randomly. This setup (\textsc{Lapt-emb}, \textsc{Reinit-random}) performs much worse than the off-the-shelf baseline in all of our experiments.

It is worth noting that \citet{artetxe-etal-2020-cross} do not necessarily suggest that freezing the main model is the \textit{optimal} language transfer method. However, it does demonstrate that for monolingual$\rightarrow$monolingual adaptation, embedding-only training is competitive with an off-the-shelf multilingual model. We see no such comparability in our experiments. We believe this is partly caused by the heterogeneity of the XLM-R embeddings, where different languages (or at least scripts) are encoded in different spaces. When new embeddings are randomly and homogeneously initialized, they fail to align with the pre-trained subspaces expected by the frozen transformer.

\paragraph{Vocab replacement efficiently specializes models}
We demonstrate that for languages inadequately covered by a pre-trained multilingual model, replacing and re-training the cross-lingual model vocabulary with a language-specific one is a computationally efficient way to create a compact model specialized for the target language(s). In our monolingual adaptation experiments, vocabulary replacement performs better than off-the-shelf XLM-R in 5/8 languages for POS tagging and 5/7 languages for NER. Only the high-resource languages of Estonian, Hebrew, and Russian seem to be adequately covered in XLM-R to outperform our specialization techniques. Language-Adaptive Pre-Training with the full (cross-lingual) XLM-R vocabulary often produces marginally better results overall, but at a much greater computational cost, and without making the model more compact in size. Further training and inference after \textsc{Lapt} will continue to suffer from the memory and compute wasted on unused vocabulary items, which constitute a large percentage of the total model parameters.

\paragraph{Script-distribution initialization rivals semantic similarity methods}
We introduced several methods for embedding re-initialization in Section~\ref{sec:reinitialization}, namely using the insight that token embeddings for XLM-R cluster by script and position within a word, then distributing new vocabulary items according to these pre-trained sub-distributions. We compare this to the \textsc{Focus} re-initialization method, which initializes new embeddings as a weighted combination of existing ones according to similarity scores from an auxiliary model.

Averaged across languages, \textsc{Focus} yields the best performance in downstream tasks by a slight margin. Within languages, it often overlaps significantly with the performance of our script-distribution methods. For very low-resource languages like Erzya, script-based methods even show a slight advantage. This seems to show that, at least in combination with \textsc{Lapt}, the majority of the benefit in re-initialization can be achieved by a method that takes the structure of the pre-trained embedding distribution into account, whether or not it uses advanced methods to precisely initialize the representations of new vocabulary items.

We do note that the advantage of \textsc{Focus} is more clear-cut when \textsc{Lapt} is conducted with transformer blocks frozen. This lends credence to the idea that \textsc{Focus} more precisely mimics the embedding distribution expected by the pre-trained transformer. However, the overall best results come when the transformer blocks are unfrozen/trainable.

\paragraph{Fully random initialization performs poorly}
Finally, our experiments demonstrate that fully random re-initialization of embeddings during vocabulary replacement leads to overall poor performance. Across \textsc{Lapt-full} experiments, random initialization performs an average of 19.4 points worse than the next-best re-initialization method, and 24.7 points worse than the off-the-shelf baseline. The poor performance of random initialization has been noted in other works such as \citet{dobler-de-melo-2023-focus}, but we emphasize that even incredibly simple methods such as \textsc{Reinit-ident} and \textsc{Reinit-script} work far better than the random baseline.

\section{Conclusion}
This work presents a systematic comparison of methods to specialize the subword vocabularies and embeddings of multilingual models for new languages. We propose simple methods for re-initializing embeddings, motivated by a qualitative exploration of the XLM-R embedding space. Our experiments show that (1) updating the encoder layers during \textsc{Lapt} is crucial for downstream performance, (2) vocabulary replacement provides a computationally-efficient method to improve task performance in low-resource languages, and (3) our re-initialization techniques employing script-wise sub-distributions perform on par with more involved similarity-based methods. We hope these findings can be built upon in future work on multilingual model specialization, with the goal of providing the best performance for under-resourced languages while also making language modeling more accessible through more manageable compute cost and model sizes.

\section*{Limitations}
One limitation of our work is the relatively narrow set of evaluation tasks available for our languages of interest. The model-adaptation techniques we compare here are most applicable to low- and medium-resource languages that are not optimally covered by pre-existing multilingual models. For most of these languages, the only standard evaluation datasets that exist are for relatively low-level tasks like Part of Speech tagging and Named Entity Recognition. Evaluation of embedding-reinitialization techniques could be improved in future work if datasets for higher-level tasks like Natural Language Inference, question answering, and paraphrase detection were curated for these under-resourced languages.

We also make several simplifying choices to maintain a feasible scope for our work. First, we conduct model adaptation from only a single base model: XLM-R. A valuable addition in future work would be to determine whether the trends we observe here generalize to other model types (i.e.~causal and seq2seq language models) and to larger model scales. Secondly, we consider only one size for newly-initialized target vocabularies (32k). Because effective per-language vocabulary allocation has been shown to be an important factor in multilingual modeling \citep[i.a.]{conneau-etal-2020-unsupervised}, investigating the dynamics of target vocabulary size during vocabulary re-initialization will be important for future work on this topic.

\section*{Acknowledgements}
We thank Ibrahim Sharaf, Anita Silva, and Peter Zuckerman for early investigation of data availability for low-resource languages. We are also gracious to Emily P. Ahn, Gina-Anne Levow, Sara Ng, and our anonymous MRL reviewers for useful feedback and discussion.

\bibliography{anthology,custom}
\bibliographystyle{acl_natbib}

\appendix

\begin{figure*}[ht]
\centering
\begin{subfigure}{0.24\textwidth}
    \includegraphics[width=\textwidth]{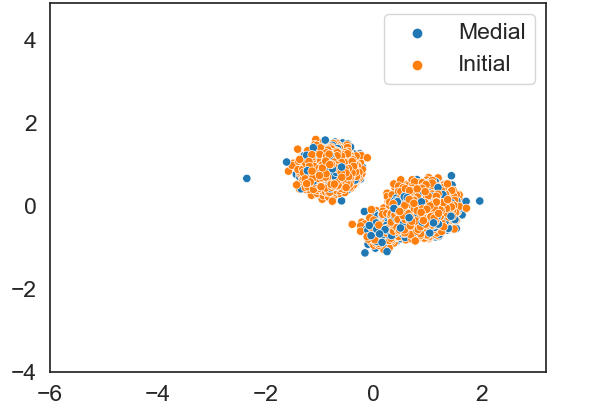}
    \caption{\textsc{script}}
    \label{fig:uralic_position_pca_script}
\end{subfigure}
\hfill
\begin{subfigure}{0.24\textwidth}
    \includegraphics[width=\textwidth]{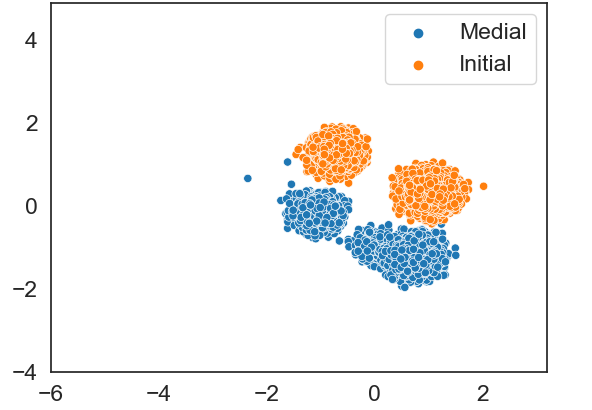}
    \caption{\textsc{script+posn}}
    \label{fig:uralic_position_pca_script+pos}
\end{subfigure}
\hfill
\begin{subfigure}{0.24\textwidth}
    \includegraphics[width=\textwidth]{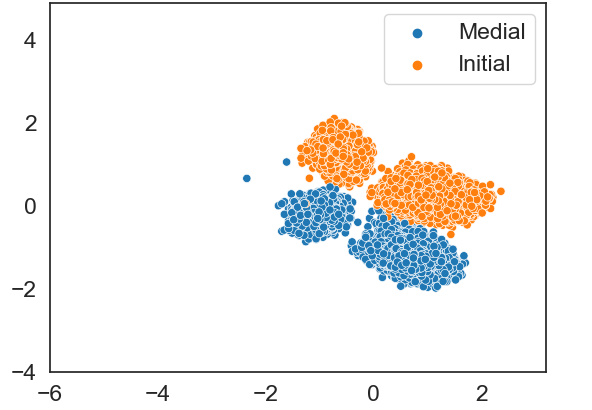}
    \caption{\textsc{script+posn+ident}}
    \label{fig:uralic_position_pca_script+pos+ident}
\end{subfigure}
\hfill
\begin{subfigure}{0.24\textwidth}
    \includegraphics[width=\textwidth]{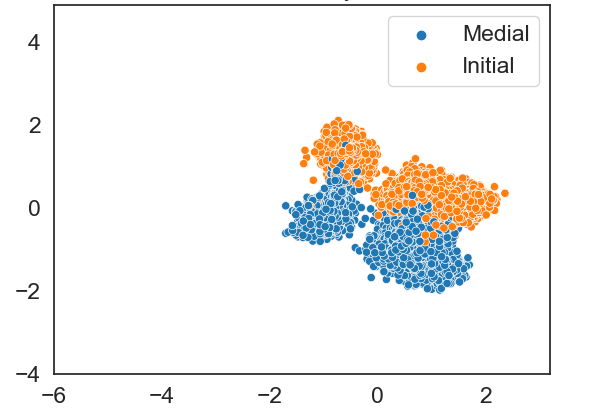}
    \caption{\textsc{focus}}
    \label{fig:uralic_position_pca_focus}
\end{subfigure}
\caption{PCA visualization of re-initialized embeddings with word-initial vs word-medial tokens highlighted. For \textsc{Reinit-script}, position-wise clustering seen in the base XLM-R embeddings (Figure~\ref{fig:base_position_pca}) is not captured. \textsc{Reinit-script+posn} and \textsc{Reinit-script+posn+ident} show expected positional clustering. \textsc{Reinit-focus} seems to allow slightly more positional overlap}
\end{figure*}

\section{Data Details}
\label{app:data}

General information about the language data used in this study can be found in Table~\ref{tab:data}. All training data used in our experiments is cleaned and deduplicated using the OpusFilter package \citep{aulamo-etal-2020-opusfilter}. For the lowest-resource languages (Erzya and Sami) we additionally filter out lines that are identified as English with a probability of 90\% or higher, since positive automatic language-identification for low-resource languages is likely not robust \citep{kreutzer-etal-2022-quality}. We additionally filter out lines composed of less than 2 tokens, lines with an average token length of greater than 16 characters, lines with tokens longer than 32 characters, and lines composed of fewer than 50\% alphabetic characters.

For POS tagging evaluation, most languages have a standard train/dev/test split curated the original Universal Dependencies dataset \citep{de-marneffe-etal-2021-universal}. Erzya, however, only has a standard train/test split. To form a dev split, we randomly sample 300 sentences from the train split. The WikiAnn dataset \citep{pan-etal-2017-cross} does not ship with standard train/dev/test splits, so we create random 85/5/10\% splits of each language for this purpose, with a minimum dev/test size of 256 and 512 sentences respectively.

\begin{table*}[ht]
    \centering
    \begin{tabular}{lccccc}
        \toprule
        Language & Code & Family & Script & XLM-R Data (GB) & \textsc{Lapt} Data (GB) \\
        \midrule
        Armenian & hy & Indo-European & Armenian & 5.5 & 1.2 \\
        Basque & eu & isolate & Latin & 2.0 & 0.35 \\
        Erzya & myv & Uralic & Cyrillic & 0 & 0.006 \\
        Estonian & et & Uralic & Latin & 6.1 & 3.0 \\
        Finnish & fi & Uralic & Latin & 54.3 & 9.1 \\
        Hebrew & he & Afro-Asiatic & Hebrew & 31.6 & 7.7 \\
        Hungarian & hu & Uralic & Latin & 58.4 & 13.0 \\
        Russian & ru & Indo-European & Cyrillic & 278.0 & 10.0 \\
        Sami & sme & Uralic & Latin & 0 & 0.004 \\
        Telugu & te & Dravidian & Telugu & 4.7 & 0.9 \\
        \bottomrule
    \end{tabular}
    \caption{Training data breakdown by language. XLM-R data is the amount of data used in the pre-training of that model. \textsc{Lapt} data is the amount used for training in our current experiments, after cleaning/deduplicating.}
    \label{tab:data}
\end{table*}

\section{Training Details}
\label{app:training}

The main details of our experimental process can be found in Section~\ref{sec:experiments}. Here we provide our choice of hyperparameters and other details relevant to reproducibility. The code used to run all experiments will be released in a later version of this paper. All models are trained and fine-tuned on Nvidia Quadro RTX 6000 GPUs using the Adam optimizer \citep{kingma_adam_2015}. 

Hyperparameters for Language-Adaptive Pre-Training (\textsc{Lapt}) can be found in Table~\ref{tab:training_hypers}. If NaN losses were encountered during training, \texttt{max\_gradient\_norm} was reduced to 0.5. For multilingual sampling during training, each language's training data is capped at approximately 2GB.

Hyperparameters for task fine-tuning on POS and NER are in Table~\ref{tab:eval_hypers}. For NER, the reported evaluation metric is entity-wise F1, meaning tokens with label \texttt{O} are ignored. In order to prevent models from learning to output only the majority class \texttt{O} during training, the loss for the \texttt{O} tokens in each batch is down-weighted to have the same influence as the tokens that actually correspond to a named entity. We cap fine-tuning training data at 32,768 sequences.

\begin{table*}[ht]
    \centering
    \begin{tabular}{lc}
        \toprule
        Hyperparameter & Value \\
        \midrule
        \texttt{mlm\_masking\_prob} & 0.15 \\
        \texttt{max\_sequence\_length} & 256 \\
        \texttt{learning\_rate} & 1e-5 \\
        \texttt{lr\_schedule} & linear \\
        \texttt{batch\_size} & 200 \\
        \texttt{max\_gradient\_norm} & 1.0 \\
        \bottomrule
    \end{tabular}
    \caption{Hyperparameters for model training (\textsc{Lapt})}
    \label{tab:training_hypers}
\end{table*}

\begin{table*}[ht]
    \centering
    \begin{tabular}{lc}
        \toprule
        Hyperparameter & Value \\
        \midrule
        \texttt{max\_sequence\_length} & 256 \\
        \texttt{learning\_rate} & 5e-6 \\
        \texttt{lr\_schedule} & constant \\
        \texttt{max\_epochs} & 64 \\
        \texttt{eval\_interval} (epochs) & 2 \\
        \texttt{patience} (epochs) & 8 (POS) / 4 (NER) \\
        \texttt{batch\_size} & 72 \\
        \texttt{max\_gradient\_norm} & 1.0 \\
        \bottomrule
    \end{tabular}
    \caption{Hyperparameters for model task fine-tuning}
    \label{tab:eval_hypers}
\end{table*}

\section{Uralic Results}
\label{app:uralic_results}
The results for multilingual adaptation to the Uralic family can be found in Tables~\ref{uralic_pos_table} and \ref{uralic_ner_table}. These results mostly follow the trends discussed in Section~\ref{sec:results} (\textsc{Lapt-emb} consistently underperforms \textsc{Lapt-full}, off-the-shelf performance is best for high-resource languages, \textsc{Lapt} with full cross-lingual vocab performs marginally better than other methods). It should be noted that for both Erzya and Hungarian, the best POS accuracy is achieved with \textsc{script+posn+ident} initialization (better even than \textsc{Lapt} with the fully cross-lingual vocabulary). Results for the very low-resource language Erzya are generally higher than with multilingual training on unrelated languages, which could suggest a benefit to training with closely-related languages. This observation does not clearly hold for Sami (the other very low-resource language), however. Note that Russian is not a Uralic language --- we include it for multilingual training in order to robustly train embeddings for the Cyrillic script, in which Erzya is written. Erzya is also spoken primarily within the Russian Federation, making loan-words likely.

\begin{table*}[ht]
    \centering
    \resizebox{0.9\textwidth}{!}{
    \begin{tabular}{llccccccc}
        \toprule
        \textsc{Lapt} & \textsc{Reinit} & Erzya & Estonian & Finnish & Hungarian & North Sami & Russian & Avg \\
        \midrule
        * & * & 56.3 $\pm$ 5.3 & 95.6 $\pm$ 0.1 & 97.5 $\pm$ 0.1 & 93.7 $\pm$ 1.5 & 71.2 $\pm$ 1.8 & 98.6 $\pm$ 0.1 & 85.9 \\
        \textsc{full} & * & 72.5 $\pm$ 2.6 & \underline{95.8 $\pm$ 0.1} & \underline{97.7 $\pm$ 0.2} & 94.1 $\pm$ 1.9 & \underline{82.9 $\pm$ 0.4} & \underline{98.6 $\pm$ 0.04} & \underline{90.3} \\
        \midrule
        \textsc{full} & \textsc{focus+ident} & \textbf{73.8 $\pm$ 2.7} & \textbf{95.3 $\pm$ 0.2} & \textbf{97.2 $\pm$ 0.1} & 92.5 $\pm$ 1.6 & \textbf{80.1 $\pm$ 1.4} & \textbf{98.4 $\pm$ 0.04} & \textbf{89.6} \\
        \textsc{full} & \textsc{script+posn+ident} & \underline{\textbf{73.0 $\pm$ 1.4}} & 94.7 $\pm$ 0.3 & 96.6 $\pm$ 0.1 & \underline{\textbf{94.8 $\pm$ 0.7}} & \textbf{78.0 $\pm$ 2.3} & \textbf{98.4 $\pm$ 0.01} & 89.3 \\
        \textsc{full} & \textsc{script+ident} & 67.7 $\pm$ 11.0 & 94.3 $\pm$ 0.3 & 96.4 $\pm$ 0.1 & \textbf{94.7 $\pm$ 0.7} & \textbf{78.8 $\pm$ 2.2} & \textbf{98.4 $\pm$ 0.03} & 88.4 \\
        \textsc{full} & \textsc{script+posn} & 71.2 $\pm$ 2.7 & 88.7 $\pm$ 0.4 & 90.6 $\pm$ 0.1 & 86.8 $\pm$ 0.4 & 72.9 $\pm$ 2.0 & 97.2 $\pm$ 0.02 & 84.7 \\
        \textsc{full} & \textsc{script} & 65.9 $\pm$ 4.6 & 85.6 $\pm$ 1.3 & 89.1 $\pm$ 0.3 & 85.2 $\pm$ 0.2 & 73.5 $\pm$ 1.6 & 96.9 $\pm$ 0.05 & 82.7 \\
        \textsc{full} & \textsc{ident} & 59.8 $\pm$ 1.2 & 92.2 $\pm$ 0.03 & 95.2 $\pm$ 0.04 & 91.8 $\pm$ 2.8 & 68.9 $\pm$ 0.9 & 98.2 $\pm$ 0.03 & 84.3 \\
        \textsc{full} & \textsc{random} & 53.7 $\pm$ 3.2 & 71.9 $\pm$ 0.6 & 73.1 $\pm$ 0.2 & 59.6 $\pm$ 1.6 & 63.9 $\pm$ 0.9 & 84.9 $\pm$ 1.9 & 67.8 \\
        \midrule
        \textsc{emb} & \textsc{focus+ident} & \textbf{66.3 $\pm$ 1.2} & \textbf{94.7 $\pm$ 0.1} & \textbf{96.8 $\pm$ 0.2} & \textbf{94.2 $\pm$ 0.8} & \textbf{73.3 $\pm$ 1.6} & \textbf{98.4 $\pm$ 0.05} & \textbf{87.3} \\
        \textsc{emb} & \textsc{script+posn+ident} & 64.2 $\pm$ 2.8 & 93.0 $\pm$ 0.1 & 95.5 $\pm$ 0.03 & \textbf{93.6 $\pm$ 0.8} & \textbf{72.7 $\pm$ 2.6} & 98.3 $\pm$ 0.05 & 86.2 \\
        \textsc{emb} & \textsc{script+ident} & 55.8 $\pm$ 4.1 & 92.8 $\pm$ 0.2 & 95.4 $\pm$ 0.04 & 92.3 $\pm$ 1.6 & 69.8 $\pm$ 1.6 & 98.3 $\pm$ 0.04 & 84.1 \\
        \textsc{emb} & \textsc{script+posn} & 54.5 $\pm$ 4.3 & 74.2 $\pm$ 0.8 & 79.5 $\pm$ 0.7 & 62.1 $\pm$ 2.6 & 65.2 $\pm$ 2.0 & 94.8 $\pm$ 0.4 & 71.7 \\
        \textsc{emb} & \textsc{script} & 48.7 $\pm$ 0.04 & 56.9 $\pm$ 15.6 & 71.6 $\pm$ 3.2 & 54.3 $\pm$ 4.4 & 58.0 $\pm$ 1.7 & 91.4 $\pm$ 1.8 & 63.5 \\
        \textsc{emb} & \textsc{ident} & 49.2 $\pm$ 1.7 & 90.6 $\pm$ 0.4 & 94.4 $\pm$ 0.03 & 84.8 $\pm$ 2.9 & 64.7 $\pm$ 1.3 & 97.9 $\pm$ 0.1 & 80.3 \\
        \textsc{emb} & \textsc{random} & 48.6 $\pm$ 0.2 & 64.5 $\pm$ 4.1 & 66.4 $\pm$ 1.2 & 43.6 $\pm$ 0.1 & 45.8 $\pm$ 4.2 & 84.0 $\pm$ 1.4 & 58.8 \\
        \bottomrule
    \end{tabular}}
    \caption{Uralic family multilingual \textsc{Lapt}: POS tagging accuracy after fine-tuning}
    \label{uralic_pos_table}
\end{table*}

\begin{table*}[ht]
    \centering
    \resizebox{0.9\textwidth}{!}{
    \begin{tabular}{llcccccc}
        \toprule
        \textsc{Lapt} & \textsc{Reinit} & Erzya & Estonian & Finnish & Hungarian & Russian & Avg \\
        \midrule
        * & * & 89.5 $\pm$ 0.6 & 93.3 $\pm$ 0.2 & 90.7 $\pm$ 0.1 & \underline{92.4 $\pm$ 0.1} & \underline{90.9 $\pm$ 0.2} & 91.4 \\
        \textsc{full} & * & \underline{90.5 $\pm$ 0.5} & \underline{93.8 $\pm$ 0.2} & \underline{91.0 $\pm$ 0.2} & 92.4 $\pm$ 0.3 & 91.0 $\pm$ 0.2 & \underline{91.8} \\
        \midrule
        \textsc{full} & \textsc{focus+ident} & \textbf{89.4 $\pm$ 1.7} & \textbf{92.5 $\pm$ 0.1} & \textbf{89.8 $\pm$ 0.2} & \textbf{91.2 $\pm$ 0.4} & \textbf{90.4 $\pm$ 0.1} & \textbf{90.7} \\
        \textsc{full} & \textsc{script+posn+ident} & 88.7 $\pm$ 0.5 & 92.2 $\pm$ 0.4 & 89.2 $\pm$ 0.2 & 90.9 $\pm$ 0.2 & 90.1 $\pm$ 0.1 & 90.2 \\
        \textsc{full} & \textsc{script+ident} & \textbf{89.3 $\pm$ 0.4} & \textbf{92.7 $\pm$ 0.3} & 89.2 $\pm$ 0.4 & \textbf{91.3 $\pm$ 0.1} & 90.0 $\pm$ 0.2 & 90.5 \\
        \textsc{full} & \textsc{script+posn} & \textbf{89.5 $\pm$ 1.0} & 87.9 $\pm$ 0.2 & 84.2 $\pm$ 0.3 & 86.3 $\pm$ 0.3 & 86.2 $\pm$ 0.2 & 86.8 \\
        \textsc{full} & \textsc{script} & \textbf{88.9 $\pm$ 0.8} & 87.5 $\pm$ 0.3 & 83.3 $\pm$ 0.1 & 86.3 $\pm$ 0.2 & 85.5 $\pm$ 0.1 & 86.3 \\
        \textsc{full} & \textsc{ident} & 81.1 $\pm$ 0.8 & 91.6 $\pm$ 0.1 & 88.2 $\pm$ 0.2 & 90.7 $\pm$ 0.3 & 89.6 $\pm$ 0.1 & 88.2 \\
        \textsc{full} & \textsc{random} & 73.7 $\pm$ 2.7 & 53.1 $\pm$ 30.7 & 0.0 $\pm$ 0.0 & 32.9 $\pm$ 33.0 & 65.1 $\pm$ 2.2 & 45.0 \\
        \midrule
        \textsc{emb} & \textsc{focus+ident} & \textbf{88.6 $\pm$ 0.6} & \textbf{92.4 $\pm$ 0.3} & \textbf{89.6 $\pm$ 0.1} & \textbf{91.1 $\pm$ 0.1} & \textbf{90.0 $\pm$ 0.1} & \textbf{90.3} \\
        \textsc{emb} & \textsc{script+posn+ident} & 86.6 $\pm$ 1.1 & 91.4 $\pm$ 0.2 & 88.8 $\pm$ 0.3 & 90.5 $\pm$ 0.2 & 89.9 $\pm$ 0.1 & 89.4 \\
        \textsc{emb} & \textsc{script+ident} & 87.0 $\pm$ 1.3 & 91.8 $\pm$ 0.1 & 88.6 $\pm$ 0.3 & \textbf{91.0 $\pm$ 0.2} & 89.6 $\pm$ 0.2 & 89.6 \\
        \textsc{emb} & \textsc{script+posn} & 85.0 $\pm$ 1.2 & 84.2 $\pm$ 0.4 & 78.1 $\pm$ 0.3 & 81.9 $\pm$ 0.5 & 82.1 $\pm$ 0.2 & 82.3 \\
        \textsc{emb} & \textsc{script} & 82.9 $\pm$ 2.6 & 82.4 $\pm$ 1.3 & 72.5 $\pm$ 1.3 & 80.7 $\pm$ 0.4 & 79.0 $\pm$ 0.2 & 79.5 \\
        \textsc{emb} & \textsc{ident} & 71.0 $\pm$ 4.4 & 90.1 $\pm$ 0.3 & 87.0 $\pm$ 0.4 & 89.9 $\pm$ 0.2 & 88.7 $\pm$ 0.1 & 85.3 \\
        \textsc{emb} & \textsc{random} & 64.9 $\pm$ 1.9 & 0.0 $\pm$ 0.0 & 13.6 $\pm$ 23.5 & 0.0 $\pm$ 0.0 & 54.4 $\pm$ 2.2 & 26.6 \\
        \bottomrule
    \end{tabular}}
    \caption{Uralic family multilingual \textsc{Lapt}: entity-wise NER F1 score after fine-tuning. A score of 0.0 results from the model learning to output only class \texttt{O} (not a named entity) which is the majority class. Sami does not have enough NER data for fine-tuning}
    \label{uralic_ner_table}
\end{table*}

\end{document}